\documentclass[conference]{IEEEtran}
\IEEEoverridecommandlockouts
\usepackage{cite}
\usepackage[utf8]{inputenc}
\usepackage[english]{babel}
\usepackage{multirow}
\usepackage{microtype}
\usepackage{tabularx}
\usepackage{amsmath,amssymb,amsfonts}
\usepackage{algorithmic}
\usepackage{graphicx}
\usepackage{textcomp}
\usepackage{xcolor}
\begin{document}

\title{Leveraging Machine Learning to Detect Data Curation Activities}

% \author{\IEEEauthorblockN{Sara Lafia}
% \IEEEauthorblockA{\textit{ICPSR} \\
% \textit{University of Michigan}\\
% Ann Arbor, MI, USA \\
% slafia@umich.edu}
% \and
% \IEEEauthorblockN{Andrea Thomer}
% \IEEEauthorblockA{\textit{School of Information} \\
% \textit{University of Michigan}\\
% Ann Arbor, MI, USA \\
% athomer@umich.edu}
% \and
% \IEEEauthorblockN{David Bleckley}
% \IEEEauthorblockA{\textit{ICPSR} \\
% \textit{University of Michigan}\\
% Ann Arbor, MI, USA \\
% dbleckle@umich.edu}
% \and
% \IEEEauthorblockN{Dharma Akmon}
% \IEEEauthorblockA{\textit{ICPSR} \\
% \textit{University of Michigan}\\
% Ann Arbor, MI, USA \\
% dharmrae@umich.edu}
% \and
% \IEEEauthorblockN{Libby Hemphill}
% \IEEEauthorblockA{\textit{ICPSR/School of Information} \\
% \textit{University of Michigan}\\
% Ann Arbor, MI, USA \\
% libbyh@umich.edu}
% }

\author{%
  \IEEEauthorblockN{%
    Sara Lafia\IEEEauthorrefmark{1},
    Andrea Thomer\IEEEauthorrefmark{2},
    David Bleckley\IEEEauthorrefmark{1},
    Dharma Akmon\IEEEauthorrefmark{1}, and
    Libby Hemphill\IEEEauthorrefmark{1}\IEEEauthorrefmark{2}%
  }%
  \IEEEauthorblockA{\IEEEauthorrefmark{1} \textit{ICPSR, University of Michigan, Ann Arbor, MI, USA}}%
  \IEEEauthorblockA{\IEEEauthorrefmark{2} \textit{School of Information, University of Michigan, Ann Arbor, MI, USA }}%
  \IEEEauthorblockA{Email: slafia@umich.edu, athomer@umich.edu, dbleckle@umich.edu, dharmrae@umich.edu, libbyh@umich.edu}%
}

\maketitle

\begin{abstract}
This paper describes a machine learning approach for annotating and analyzing data curation work logs at ICPSR, a large social sciences data archive. The systems we studied track curation work and coordinate team decision-making at ICPSR. Repository staff use these systems to organize, prioritize, and document curation work done on datasets, making them promising resources for studying curation work and its impact on data reuse, especially in combination with data usage analytics. A key challenge, however, is classifying similar activities so that they can be measured and associated with impact metrics. This paper contributes: 1) a schema of data curation activities; 2) a computational model for identifying curation actions in work log descriptions; and 3) an analysis of frequent data curation activities at ICPSR over time. We first propose a schema of data curation actions to help us analyze the impact of curation work. We then use this schema to annotate a set of data curation logs, which contain records of data transformations and project management decisions completed by repository staff. Finally, we train a text classifier to detect the frequency of curation actions in a large set of work logs. Our approach supports the analysis of curation work documented in work log systems as an important step toward studying the relationship between research data curation and data reuse.
\end{abstract}


\section{Introduction}
\label{introduction}
Data curation – the work needed to make a dataset fit-for-use over the long-term – is critical to eScience \cite{Lord2004-zr, Pennock2007-eb, Goble2008-vn, Palmer2011-ud, Johnston2017-cq}. Datasets are almost never analysis- or preservation-ready upon initial collection, and extensive pre-processing, cleaning, transformation, documentation, and preservation actions are required to support data’s usability, sharing, and management over time. However, despite extensive development of data curation best practices, the impacts that specific curatorial activities have on data use and reuse are unclear. We use supervised machine learning techniques to analyze a corpus of Jira tickets documenting data curation activities at a large social sciences data archive, the Inter-university Consortium for Political and Social Research (ICPSR) at the University of Michigan. We ask: 1) What are the main tasks of curatorial work at a large scale data science repository?; 2) How do curators document their curation work?; and 3) How do curatorial actions vary across projects and requests?

We define \textit{curatorial actions} as the specific steps taken to improve data products’ fitness-for-use or preservation readiness. These include tasks such as data normalization; creating and improving metadata and other documentation; the application of controlled vocabularies or standards; and so on. These tasks vary and depend on the type of data being curated; the scope and focus of the organization doing the curation; and the designated community the curators seek to serve \cite{Pienta2010-bl, Daniels2012-pj, Yakel2019-na}. Chao, Cragin and Palmer \cite{Chao2015-bq} derive a typology of data curation concepts, activities, and terms through a qualitative study of earth science researchers and show how the typology can support cost analysis of different curatorial activities. However, further work is needed to examine the efficacy of these tasks across different data types and to empirically demonstrate the costs and benefits of the activities to a repository and a user community.

The use of project management systems (e.g., Jira, Asana, Trello) in large-scale curatorial settings presents us with a rich potential data source to study curatorial actions. Through routine use of systems such as Jira, data curators at ICPSR generate a corpus documenting data curation work and the frequency of specific curatorial activities. These data offer an abundant source of information about the impact and efficacy of different curatorial processes, and, given their scale and structure, computational methods are useful for analyzing them. ICPSR adopted Jira to facilitate existing work practices, and the content staff generated in the system documented the “articulation work” \cite{Strauss1988-vy} of curation. Their comments make visible the details of curation tasks and hint at the related organizational processes. Analysis of these work logs is important in revealing the often hidden ways in which people, data, and data processing algorithms are brought together to produce refined datasets ready for analysis. These work logs can thereby complicate accounts of data production as a straightforward pipeline, and instead show the importance of the “humans-in-the-loop”. Additionally, identification and analysis of curatorial activities is an important first step in showing the long-term impact and value of these activities.

The eScience community has called for improved data curation tools and analysis to support new ways of doing science \cite{Hey2009-lk}. Digital data archives like ICPSR are critical components of knowledge infrastructures \cite{Borgman2019-ha}. As sites of scientific data curation, they play a key role in managing, preserving, and disseminating knowledge. ICPSR is a well-established social science archive, curating data at scale across a breadth of sub-disciplines ranging from criminal justice to early child care and education. Over 3,000 studies were deposited at ICPSR from 2016 to 2020 \cite{ICPSR-2020}. Human effort to curate data products is often rendered invisible to those outside of such archives out of an impulse to create a “pristine” dataset that deliberately obscures the curator’s mark on it \cite{Plantin2019-ba}. However, curators’ specialized disciplinary knowledge and labor is essential to the enterprise of data curation \cite{Cragin2010-ei}. Our research values curators’ specialized knowledge by making their labor explicit and measurable.

This paper describes a study of Jira work logs to better understand common curation activities as a first step toward connecting curation activities with data reuse and impact. We recognize that, like any documentation system, ICPSR’s curation work logs are incomplete and vary in specificity; even with the adoption of systems like Jira, some curation work goes undocumented or is obscured by the level of recorded detail. Therefore, our results tell a partial story about curation activities; we likely underestimate how often actions are performed. However, these estimates serve as a baseline for defining categories of curation activity and measuring the frequency and effort time of various categories of curatorial actions performed on social science datasets. 

We identified eight main categories of tasks that are frequently recorded in curation work logs, two of which were \textit{non-curation} or \textit{other} kinds of activities (e.g., creating training materials, attending staff meetings). Excluding these, \textit{quality checks}, \textit{initial review and planning}, and \textit{data transformation} were the most frequent and time consuming curatorial activities recorded across all of the studies in our analysis. On average, curators spent more time on studies assigned higher levels of curation, confirming that applying more intensive sets of curation actions requires more staff time. 

Our analysis covers a period of transition as ICPSR standardized its curation work; we observed changes in curation actions over this time across levels of curation and between ICPSR archives. The average amount of time spent curating studies has been decreasing since 2017, signaling possible increases in efficiency in ICPSR’s curation practices. For example, fewer instances of \textit{data transformation} were performed over time on deposits in topical archives while \textit{initial review and planning} became more common; in the ICPSR General Archive however, the frequency of \textit{data transformation} remained constant while \textit{initial review and planning} were performed more often. For intensively curated studies, \textit{initial review and planning} was recorded more frequently than it was for non-intensive studies; \textit{communication} was also recorded more frequently for intensively curated studies, although this decreased over time. To analyze data curation work, we contribute:  

\begin{enumerate}
    \item  a schema of data curation activities; 
    \item  a computational model for identifying curation actions in work log descriptions; and
    \item an analysis of the frequency and effort associated with particular curation activities.
\end{enumerate}

\section{Background}
\label{background}

\subsection{Data curation activities}
\label{data-curation-activities}
Data curation plays a critical role in enabling accessibility, discovery, re-use, preservation, and data sharing \cite{Tenopir2011-an}. Yet, as several researchers in this area have noted \cite{Carlson2012-lb, Pham2018-ob}, the term is often ill-defined with little explication of the specific activities that comprise “data curation.” In an effort to clarify what is meant by “data curation,” a number of researchers and practitioners have defined it by describing what data curation enables. They define data curation as “the management and promotion of data from the point of its creation, ensuring the fitness of data for contemporary purposes, and making data available for discovery and re-use” \cite{Carpenter2004-wg, Carlson2012-lb}; as “actions taken on data sets at any stage of their existence...that enhance their use or reuse value” \cite{Darch2020-rx}; and as “the process of managing research data throughout its lifecycle for long-term availability and reusability” \cite{Lee2017-be}. Johnston and colleagues \cite{Johnston2018-gy} call attention to repository curators as the key actors in data curation, describing it as “work and actions taken by curators of a data repository in order to provide meaningful and enduring access to data.” 

To better get at what, precisely, this work and these actions are, several studies have developed lists and taxonomies of specific curation tasks; however, we have found that existing vocabularies of curatorial actions are not readily applicable to workflow documentation at ICPSR. The language of curation documented in the literature differs from the terminology that curators we studied used in practice. For instance, Johnston and colleagues \cite{Johnston2018-gy} create a ranked list of 47 individual curation activities based on focus groups with researchers to identify the curation activities they most valued, resulting in “documentation” (ranked first); “chain of custody” (second); and “secure storage” (third). Many terms describe automated actions applied to all data in a large-scale repository like ICPSR (e.g., “terms of use,” “use analytics,” and “secure storage”) and, hence, are not useful in studying the day-to-day work of human data curators. Furthermore, while many of the activity definitions (e.g., “Use metadata to link the data set to related publications, dissertations, and/or projects that provide added context for how the data were generated and why”), would be readily recognized by the curators we studied, the terms (e.g., “contextualizing”) are not consistent with the language ICPSR curators use to characterize their work. 

The Data Practices and Curation Vocabulary (DPCVocab) \cite{Chao2015-bq} was derived from interviews with data managers in the earth sciences. The DPCVocab links data curation practices with products, curation workflows, and curation roles and functions including stakeholders and stewards. One of the key aims was to create a practical vocabulary for curators that could be used in the curation of datasets from all fields of science. Within the “curatorial actions and functions” category, the DPCVocab includes categories of activities that are synonymous with the function of an archive like ICPSR (e.g., ingest, representation, provenance management, data storage, policies, and preservation). However, it places the hands-on work with data that is often part of curation work (e.g., validating data) as “research data practices” performed primarily by the scientists who created the data. Lee and Stvilia \cite{Lee2017-be} identify 14 research data activities and actions in an institutional repository, which span communication about data curation needs, managing and sharing data, ensuring data accessibility, and re-evaluating data for long term preservation. But again, these do not align with ICPSR workflows. Finally, the RDA/TDWG Curatorial Metadata and Attribution Model \cite{Thessen2019-ay} proposes an abstract data model for describing and citing curatorial work; rather than prescribing specific curatorial tasks, it allows curators to receive credit for work. However, this model is meant to be applied in conjunction with an existing taxonomy of curatorial work, and does not outline curatorial tasks.

This breadth of vocabularies regarding curatorial actions reflects the diversity of data curation contexts and domains, and points to a need for further research on the nature of data curation work. These vocabularies were developed through interview-based methods, not by examining existing documentation; therefore they are not ideal for supporting text extraction and classification of curatorial work logs. Though there appear to be high-level commonalities between these vocabularies, more specific accounts of data curation are needed to render this important work visible, and to support institutions in assessing the efficacy of their own curatorial pipelines. In the section that follows, we outline data curation pipelines at ICPSR so as to foreground the development of our own data curation taxonomy.

\subsection{Data curation at ICPSR}
\label{data-curation-at-ICPSR}
Prior studies suggest that the adoption of standards can make implicit institutional practices, like data curation, more explicit so that efforts and funds can be prioritized \cite{Mayernik2016-pa}. In 2017, ICPSR took a significant step toward standardizing curation work by centralizing curation staff into an organizational unit. ICPSR’s entire repository of social science data is organized around several thematic collections or “topical archives” with each archive corresponding to a particular social science research audience (e.g., the National Addiction \& HIV Data Archive and the National Archive of Computerized Data on Aging). Prior to the 2017 reorganization, each topical archive within ICPSR employed its own team of curators, leading to unique, project-specific approaches to curating data. Following the reorganization, ICPSR also established a set of written curation standards that grouped specific curatorial actions and outputs into three different curation levels that vary with respect to the amount, intensiveness, and complexity of effort required as well as the end product. 

All studies deposited at ICPSR receive a base level of curation called “Level 1”: curators conduct a disclosure risk review and create a study webpage with subject terms and relevant study information including a description, title, authors, and relevant notes. Level 1 curation also includes an ICPSR codebook and data files for all major statistical software packages. “Level 2” builds on Level 1, further improving the usability of data by ensuring missing values are identified and documented, acronyms and abbreviations are spelled out, spelling is checked and corrected, and labels are checked for completeness and readability. “Level 3” is ICPSR’s most intensive level of curation; it develops custom documentation for the data and adds survey question text to variables so that they can be indexed and searched. This level is also applied to non-tabular or non-numeric data such as GIS and qualitative data. In developing our curatorial schema in Section~\ref{annotation-schema}, we include actions that are performed on every study across curation levels but which vary in intensiveness; for example, disclosure risk review is applied to all studies but Level 3 curation tends to involve more steps, such as reviewing all variables and survey question text for disclosive information. 

\subsection{Recording curation activities}
\label{recording-curation-activities}
In an additional move to systematize the curation workflow, ICPSR adopted Jira to document, prioritize, and communicate curation work. Jira is a highly-customizable web-based tool, most commonly used by software developers to plan, track, and release software. Previous studies have performed text mining on issue tracking systems to glean behavioral insights into the communication styles of developers \cite{Ortu2015-da}. Systems like Jira offer new ways to document and analyze the data curation process. Given the need to rationalize return on investments for data curation \cite{Parr2019-uz}, we are interested in studying work management systems to better understand the effects of particular curation activities on data.  

At ICPSR a curation “ticket” or “issue” is synonymous with a “curation request”; therefore, we use the terms interchangeably. Jira tickets are the primary means for making the request to curate a study (often made up of multiple files); the tickets list the necessary curation tasks, communicate about curation, track time-stamped milestones, and document work from start to resolution (i.e., study release). 

\section{Method}
\label{method}
\begin{figure}[ht]
\includegraphics[width=\linewidth]{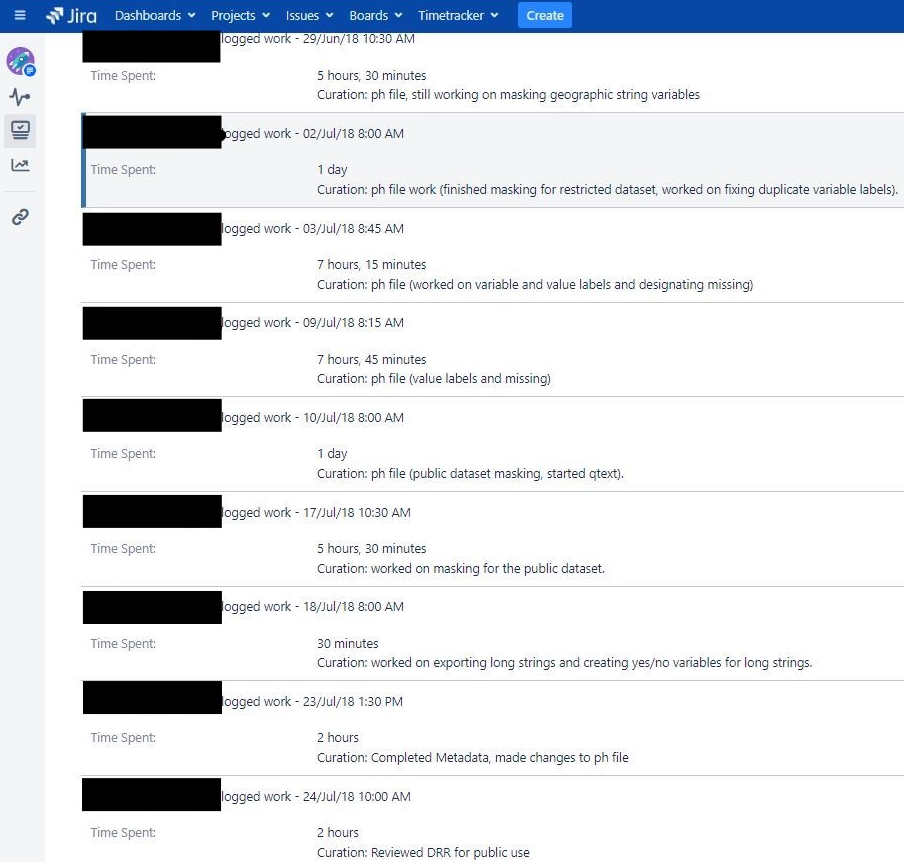}
\caption{Anonymized view of a curator’s work logged in ICPSR’s Jira ticketing system}
\label{fig:figure1}
\end{figure}

Jira tickets contain work logs, which describe the curation work performed on deposited data (Figure~\ref{fig:figure1}). To identify curation actions in the work logs, we first developed a schema of curation activities and then manually labeled a subset of work logs according to the type of data curation activity that each described. We trained a supervised classifier using the labeled data to predict curation actions in unlabeled work logs. We compared results from two classification models (Complement Naive Bayes and Stochastic Gradient Descent) to a baseline model that applied labels proportionally. Below, we describe our corpus, our pre-processing workflow, our annotation schema and labeling method, and our classifier in more depth.

\subsection{Jira ticket corpus and preprocessing}
\label{jira-ticket-corpus}
We analyzed a corpus of Jira tickets created between February 2017 and December 2019. The start date coincided with ICPSR’s adoption of the Jira system; we omitted tickets created from 2020 onward as curation was still in progress for many of these at the time of writing. We included only tickets with at least one work log entry written by ICPSR staff during curation. The corpus of 669 Jira tickets corresponded to 566 unique studies.  

We deidentified work log text by replacing curators’ names with linked anonymous identifiers. We segmented the work log descriptions into short fragments and applied term frequency-inverse document frequency \cite{Sparck-jones1972-tq} to identify important curatorial phrases. This preliminary analysis suggested that the description of curation activities within the work logs was not consistent. For example, descriptions of \textit{quality checks} included phrases such as “self-checks”, “1QC,” and “addressing identified issues.” Many curation actions were also described with generic phrases; for example, “wrapping up study” and “running scripts” imply activities that include \textit{quality checks} but are not explicit enough to classify as such without more context. Work logs tended to overgeneralize work so that broad categories rather than specific actions were captured. Descriptions of work also varied in complexity, ranging in length from 1 to 204 words. 

\subsection{Manual annotation schema}
\label{annotation-schema}
To account for the variety in the work log descriptions, we developed a schema of curation activities. We focused on curatorial actions that vary in application across studies by curation level and relative amounts of time spent. More background and definitions for ICPSR’s curation levels are given in Section~\ref{data-curation-at-ICPSR}. We first consulted with ICPSR internal documentation and curation supervisors to identify an exhaustive list of curation actions. We then sorted these actions according to how frequently they were performed (i.e., across all studies vs. as needed) and how variable time spent on them was (i.e., a consistent amount of time vs. dependent on the dataset). We used this initial set of frequent actions with high variability in our first annotation attempts; we then iteratively revised the schema as a team until we settled on eight comprehensive, mutually exclusive categories of curatorial actions (Table~\ref{table:definitions}). 

The first four actions listed occur in succession; \textit{quality checks} tend to happen at the end of other actions. \textit{Communication} for study is done as needed throughout the curation process. While we acknowledge similarities between the two terms, we distinguished between \textit{documentation} and study \textit{metadata} as the standalone human-readable descriptive files (codebooks, record layouts, questionnaires, technical reports, etc.) and machine-readable descriptive information, respectively. We included \textit{other} as a category to capture curation-related actions outside of our designated categories. We also identified \textit{non-curation} actions, which we removed from our analysis as reported in Section~\ref{results}.

\begin{table}[t]
\caption{Summary of curatorial actions in annotation schema}
\centering
\begin{tabularx}{\columnwidth}{|l|X|}
\hline
\textbf{Curatorial Action}           & \textbf{Examples}                                                                                                                                                             \\ 
\hline
\textit{Initial review and planning} & 
Look at deposited files, determine curation work needed, compose processing plan, create processing history syntax                                                                                              \\ 
\hline
\textit{Data transformation}         & Locate identifiers, revise or add variable/value labels, designate or fix missing values, reorder/standardize/convert variables, create variable-level metadata, collapse categories for disclosure                                             \\ 
\hline
\textit{Metadata}          & Draft or revise study description, copy metadata from deposit system, update collection dates based on dataset, create survey question text, describe variable level labels                                                      \\ 
\hline
\textit{Documentation}               & Create a codebook, document major changes or issues with the data, compile documentation archived by the data producer                                                        \\ 
\hline
\textit{Quality checks}              & Check all files and metadata for completeness, adherence to standards, alignment with JIRA request after all data and documentation curation is complete (Self QC, 1QC, 2QC)  \\ 
\hline
\textit{Communication}     & Discuss study with project manager, consult supervisor on curation standards for study, check how to handle specific variables                                                \\ 
\hline
\textit{Other}                       & Compile folders for study, ambiguous or overly-general curation work                                                                                                         \\ 
\hline
\textit{Non-curation}                & Staff meetings, timesheets, administrative work                                                                                                                               \\
\hline
\end{tabularx}
\label{table:definitions}
\end{table}

We uploaded a proportional random sample of Jira tickets across curation levels 1-3 to a web-based annotation tool, BRAT \cite{Stenetorp2012-bw}. We applied the schema to manually annotate a set of 789 labeled curation actions from 10 randomly-selected tickets to use as training data for text classification.  

\subsection{Computational model for text classification}
\label{classifier}
\begin{figure}[t]
\includegraphics[width=\linewidth]{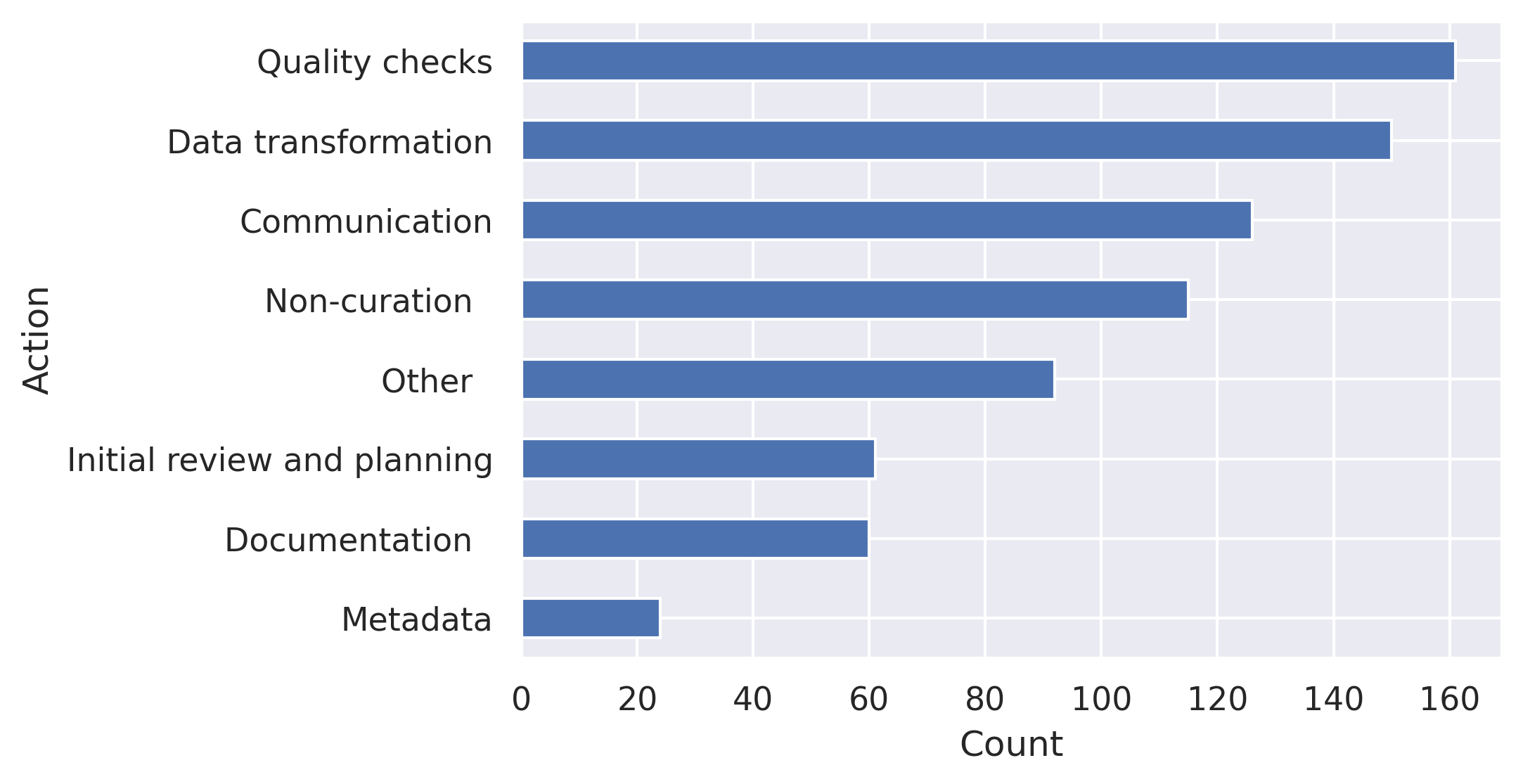}
\caption{Distribution of labeled curation actions for each schema class}
\label{fig:figure2}
\end{figure}

Our objective was to classify curation actions in unlabeled curation work logs. Methods like supervised classifiers reduce the labor needed to detect and distinguish specialized curatorial language in short, unstructured text \cite{Bird2009-vi, Hemphill2020-mv}. We chose a supervised approach that leveraged the manual annotations to train a machine classifier to recognize curation activities. Our labeled data had many instances of \textit{quality checks} and fewer instances of study level \textit{metadata} (Figure~\ref{fig:figure2}). To train a classifier to predict curation actions, we split the labels generated in BRAT into 80\% training and 20\% testing datasets. We removed stopwords from the labeled data and constructed ngrams of lengths 1 and 2. We then stored the labeled data as a document term matrix for retrieval with our classifier.  

We trained a stratified baseline dummy model, which we compared to two other supervised approaches appropriate for classifying large amounts of text: Complement Naive Bayes (NB) and a linear model, stochastic gradient descent (SGD). Complement NB is suited for text classification tasks with imbalanced class distributions \cite{Rennie2003-bb} while SGD has been found to perform well on large, sparse text datasets \cite{Zhang2004-jb}. Using these models, we achieved substantial gains in performance over the baseline model. We compared the performance of the classifiers with an ensemble of performance measures including accuracy, defined as the proportion of correctly predicted ground truth labels (Table ~\ref{table:metrics}). By this measure, the Complement NB classifier gave the best predictions for the test classes.

\begin{table}[t]
\centering
\caption{Comparison of supervised classifier performance scores}
\begin{tabularx}{\columnwidth}{l|X|X|X|X}
\textbf{Classifier} & \textbf{Accuracy} & \textbf{F1} & \textbf{Precision} & \textbf{Recall}     \\
\hline
\textbf{Baseline} & 0.15 & 0.14 & 0.14 & 0.14                         \\
\textbf{SGD Classifier}	& 0.73	& 0.72	& 0.74	& 0.72          \\
\textbf{Complement NB}	& 0.75	& 0.74	& 0.76	& 0.75 
\end{tabularx}
\label{table:metrics}
\end{table}

We applied the trained classifier to identify curatorial actions in the unseen Jira work logs. Work log syntax showed that curators used new line and carriage key delimiters along with line breaks and periods to separate curation actions in their work logs. We used these delimiters to segment work log descriptions into short fragments, resulting in 12,995 unseen curation sentence fragments. We then predicted a curation action for each fragment using our trained model.  

\section{Results}
\label{results}
Our schema defines the main tasks of data curation work at a large data science repository, ICPSR. We find that the coverage of the schema is sufficient to characterize and distinguish general categories of intensive curatorial actions. We use our computational model to detect curation actions in Jira tickets. We interpret the model to understand the variety in the language of curation, including sources of confusion that the classifier encountered. Finally, we analyze the output of the classifier to characterize the degree to which curatorial work varies over time with respect to data curation levels and archives within the repository.

\subsection{Defining curatorial actions}
\label{defining-curatorial-work}

\begin{figure}[b]
\includegraphics[width=\linewidth]{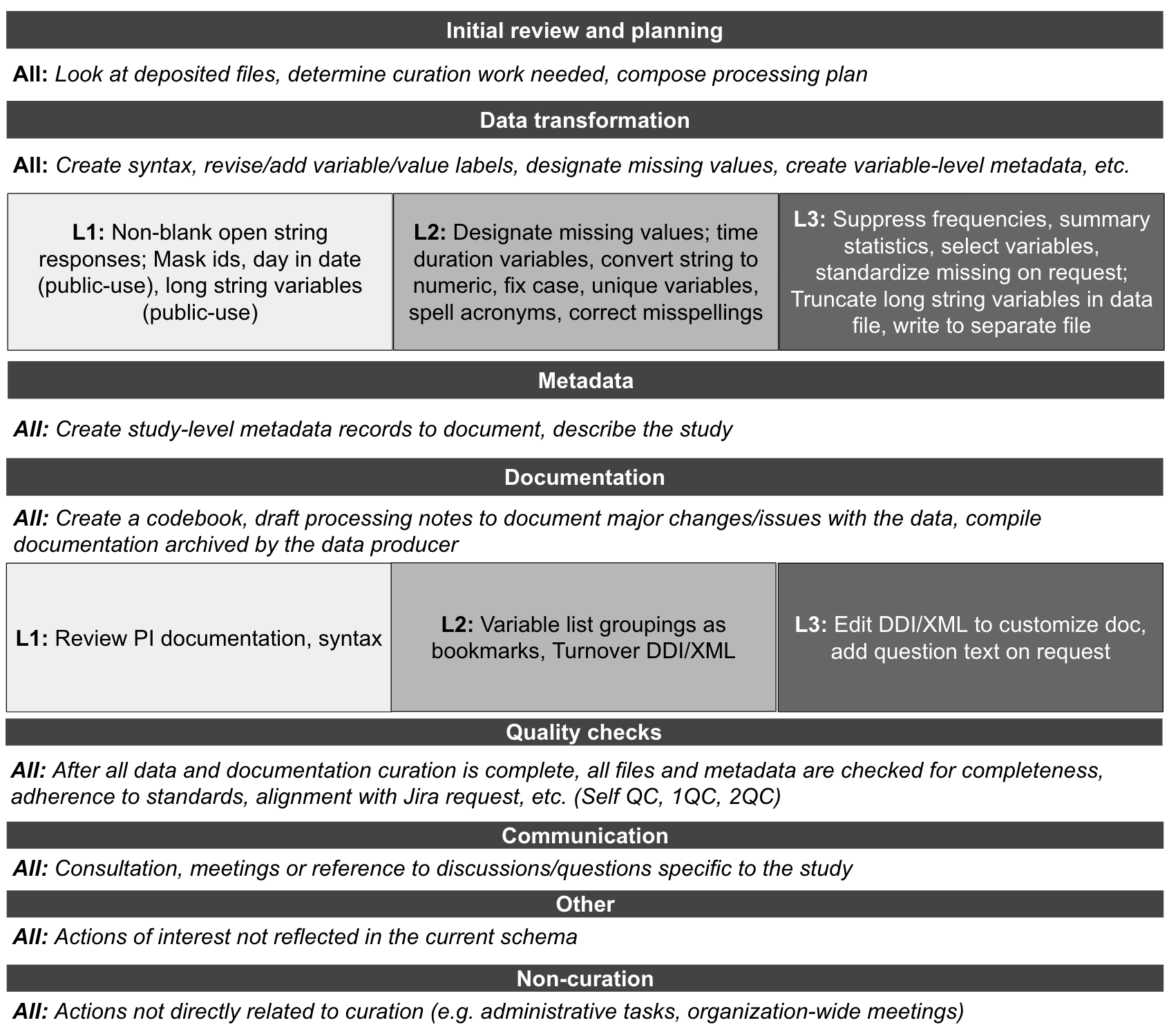}
\caption{Revised schema with activities distinguished by curation levels (L1-L3)}
\label{fig:figure3}
\end{figure}

The annotation schema of curation activities we developed (Figure~\ref{fig:figure3}) allowed us to control for variation in curator styles and changing norms in the use of Jira at ICPSR over time. Given our interest in understanding the variation in curation work across studies, we included actions that varied in frequency and effort. We initially tested annotation with a schema of 25 terms that included frequent instances of each example in Table~\ref{table:definitions} (e.g., designate missing values as a frequent instance of \textit{data transformation}); however, given the variation in detail of work log descriptions, this proved to be too granular to apply consistently. We refined the schema to focus only on the broad categories of curation actions. We also added \textit{non-curation} and \textit{other} categories so that we could differentiate usage of the Jira ticket (e.g., professional development activities); for the purposes of our analysis, these actions were not relevant. For further context on this distinction, see Section~\ref{understanding-impact-of-organizational-structure}. The initial categories of “disclosure risk remediation” and “processing history” were merged under \textit{data transformation} in the revised schema. We also added \textit{quality checks} and \textit{communication}, as we found these actions were applied across all studies but complemented the established categories rather than falling within them. 

We found that the revised schema was comprehensive enough to annotate the vast majority of work logs in Jira tickets sampled across curation levels. We only used the other category in several cases where an action was clearly supporting curation but fell outside of the established action types (e.g., “compiling folder”) or was ambiguous (e.g., “curation work”). We also found that many instances of \textit{communication} were documented alongside other actions (e.g., “asked about the content for string responses... and if any additional information could be provided”) making it difficult in some cases to differentiate discrete curation actions. In cases where curators described transforming data or revising documentation in response to a quality check, we labeled the action as a quality check even though \textit{data transformation} and \textit{documentation} may have also been relevant.

\subsection{Detecting curatorial actions}
\label{detecting-curation-activities}
Overall, our model performed well (F1 = 0.74). It revealed several categories of curation work that were straightforward to detect and others that created confusion, indicated by incorrect predictions. The classifier performed best in detecting \textit{communication}, \textit{quality checks}, \textit{non-curation}, and \textit{data transformation} actions, suggesting that the language used to describe these is internally consistent. While smaller in size, there was also complete agreement between all predictions of \textit{metadata} related actions. When the model made mistakes, it confused \textit{data transformation} with other classes. We inspected the mislabeled instances and noticed that classes with the least confusion also exhibited more homogeneity in the language used (e.g., “quality checks”). It makes sense that a bag-of-words classifier would have lower performance when the language is more diverse as it is in \textit{initial review and planning} and \textit{data transformation}. Activities that are part of these general classes of curation activity also overlap (e.g., talking to a supervisor about a \textit{data transformation}), which may explain the model’s confusion. 

\subsection{Identifying differences in curatorial actions}
\label{identifying-differences-curatorial-actions}
We describe our corpus of Jira tickets along with information about the amount of time spent on curation (Table ~\ref{table:tickets}). More time on average was spent curating Level 3 studies, supporting the idea that higher levels of curation tend to be more time intensive. Relatively less time was spent curating studies in one of ICPSR’s large topical archives – the Bureau of Justice Statistics (BJS) within the National Archive of Criminal Justice Data – compared to the ICPSR General Archive and all the other topical archives combined. The average amount of time spent curating studies has also been decreasing since 2017, signaling possible changes in curation practices or their efficiency.  

\begin{table}[t]
\centering
\caption{Description of Jira ticket corpus of curation requests}
\begin{tabularx}{\columnwidth}{|X|X|X|X|X|X|}
\hline
                                   &         & \textbf{Total \newline tickets \newline (n=669)} & \textbf{Total \newline studies \newline (n=566)} & \textbf{Average curation hours/study}  \\
                                   \hline
\multirow{3}{*}{\textbf{Curation}} & Level 1 & 221                           & 178                           & 51                                  \\
                                   & Level 2 & 229                           & 210                           & 79                                  \\
                                   & Level 3 & 219                           & 178                           & 165                                 \\
                                   \hline
\multirow{3}{*}{\textbf{Archive}}  & BJS     & 131                           & 124                           & 78                                  \\
                                   & ICPSR   & 116                           & 104                           & 105                                 \\
                                   & Other   & 422                           & 338                           & 102                                 \\
                                   \hline
\multirow{3}{*}{\textbf{Year}}     & 2017    & 133                           & 119                           & 107                                 \\
                                   & 2018    & 305                           & 276                           & 99                                  \\
                                   & 2019    & 231                           & 171                           & 88     \\
                                   \hline
\end{tabularx}
\label{table:tickets}
\end{table}

Each predicted curatorial action corresponded to an amount of time logged in a ticket. To estimate hours associated with each kind of curation activity, we summed the hours logged for each kind of curation action, which we divided by the total curation hours logged across all tickets. We report this as a percent of total work log hours; we also report the frequency of curation actions as the percent of studies containing each type of action (Table~\ref{table:actions}).

\begin{table}[t]
\centering
\caption{Studies recording curation actions and percent of hours logged across all studies}
\begin{tabularx}{\columnwidth}{|l|X|X|}
\hline
\textbf{Action}             & \textbf{Percent of studies containing action} & \textbf{Percent of total work log hours classified as action}   \\ \hline
\textit{Quality checks}              & 90.1                            & 31.6                                    \\ \hline
\textit{Initial review and planning} & 70.0                              & 14.0                                     \\ \hline
\textit{Data transformation}      & 67.6                            & 29.9                                     \\ \hline
\textit{Metadata}        & 57.7                            & 6.5                                      \\ \hline
\textit{Documentation}               & 56.2                            & 7.5                                      \\ \hline
\textit{Communication}     & 54.6                            & 7.9                                      \\ \hline
\textit{Other}                       & 40.9                            & 2.8                                      \\ \hline
\end{tabularx}
\label{table:actions}
\end{table}

\textit{Quality checks}, \textit{initial review and planning}, and \textit{data transformation} were the most recorded activities across all of the studies in our analysis. We find that these curation actions are both frequent and time consuming. Other actions, including \textit{initial review and planning}, were also recorded frequently across all studies but were not as time consuming in the aggregate. Actions like \textit{communication} were not recorded as frequently across all studies, suggesting that the Jira ticketing system is used to record work done directly to data; the substrate of data work, including \textit{documentation} and \textit{communication}, does get recorded but is not logged as frequently or for as long of a duration in aggregate as other kinds of actions, like \textit{quality checks} and \textit{data transformations}.  

\begin{figure}[ht]
\includegraphics[width=\linewidth]{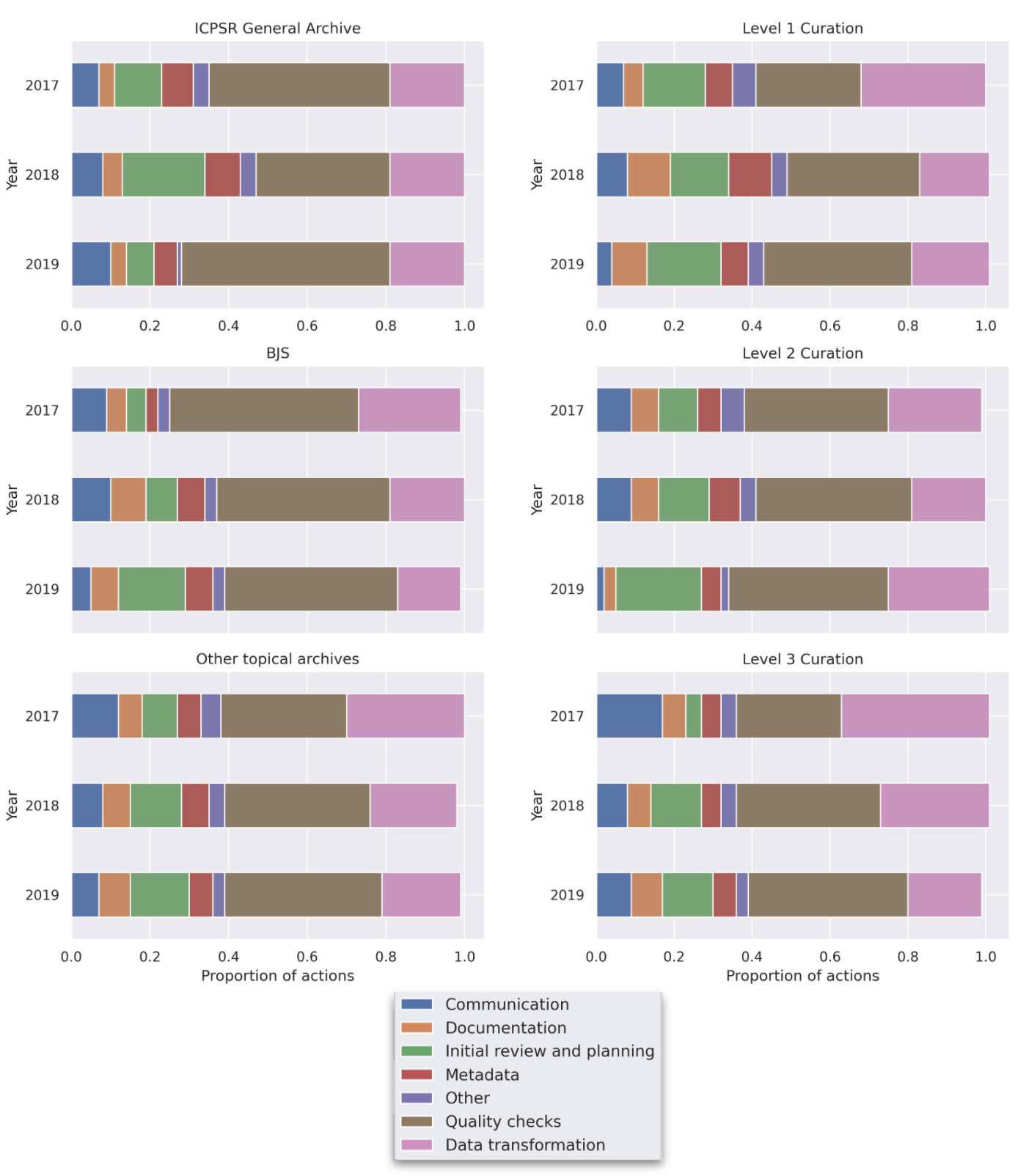}
\caption{Proportion of curation actions recorded by study curation level, archive, and year}
\label{fig:figure4}
\end{figure}

Our study covered a period of institutional transition starting in 2017. To interpret how curation actions changed over this period, we examined differences in curatorial actions between levels of curation and archives over time. Proportionally, \textit{quality checks} and \textit{data transformations} were the most frequently recorded actions across curation levels and archives (Figure~\ref{fig:figure4}). Tickets for Level 2 and 3 curation recorded more project related \textit{communication} than Level 1 curation. BJS, which is a topical archive, recorded more instances of \textit{documentation} than the ICPSR General Archive and all other topical archives (grouped under “Other topical archives”).

From 2017 to 2019, \textit{data transformation}, \textit{communication} and \textit{other} related curation actions decreased while \textit{initial review and planning} and \textit{documentation} increased. Despite this, the proportions in which curation actions are applied across studies is consistent in the aggregate across curation levels and archives for our corpus of Jira tickets. In recent years, ICPSR has been emphasizing the automation of repetitive curation tasks to prioritize value-added work that requires human expertise; ICPSR curators may be spending more time on actions like \textit{initial review} of data and \textit{quality checks} and less time on \textit{data transformation} due to this shift in priorities. Additionally, thorough \textit{initial reviews} may expedite \textit{data transformation}.

\section{Discussion}
\label{discussion}
Prior studies of curation work have primarily focused on articulating the activities that fall under the category of \textit{data transformation} \cite{Doty2014-wy, Chao2015-bq, Johnston2017-cq}. Our results surface other kinds of curation activities like \textit{quality checks} and \textit{initial review}; we additionally show that these meta-level review activities tend to be even more frequent than \textit{data transformation}. We also find that activities like \textit{metadata} creation, \textit{documentation} of curatorial work, and \textit{communication} with other stakeholders take considerable time regardless of curation level and archive. This broader portrait of curatorial work is important in developing data curation pipelines and best practices going forward. Further work is needed to understand the impact of these separate tasks on increasing the value of deposited data. Accounting for the differential impacts of curation tasks may help to define some of the value of depositing data in a trusted repository. For example, it is unclear whether individual researchers or teams preparing data for self-deposit attend to \textit{quality checks} with as much detail as professional curators. 

This study contributes a novel schema of data curation activities. Though specific to the workflows of ICPSR, we believe it has several applications beyond this. First, the method used to develop the schema may be adapted in other contexts, both to facilitate the analysis of similar collections of work logs, and also to foster a better understanding of curatorial work and workflows. Many data curation workflow models \cite{Higgins2008-ki, DDI-Lifecycle} envision curation as a set of easily identifiable, sequential, and discrete steps. The reality of work is much harder to itemize, however. In our analysis, we find examples of curators working in parallel rather than sequentially, which complicates many accounts of data curation and data science workflows (and which may explain why some instances of similar work fall into different categories). Second, the schema of curatorial work we have developed can be compared to taxonomies rooted in other contexts \cite{Chao2015-bq}, and thereby support a more nuanced understanding of data practices across disciplines.

We also introduced a machine learning classifier that detected curation activities identified in our schema. The classifier performed well in detecting the most frequent categories of curatorial actions (\textit{communication} for study, \textit{quality checks}, and \textit{data transformation}). However, no classifier is perfect, ours included. In some cases, the classifier was confused by what constituted a discrete action or distinctions between categories of complementary actions, such as \textit{data transformations} and \textit{quality checks}. An example of a mislabeled work log was “went through processing history files”, which was manually labeled as a quality check because it referred to a person reviewing the files that curators generate when editing data; however, the classifier labeled it \textit{data transformation}, likely because other work logs with “processing history files” referred to the process of generating that document instead of reviewing it. Such examples highlight compound, multi-part, or iterative decisions recorded by data workers, which are difficult to render visible without an in-depth understanding of the work context. 

\subsection{Implications for future work}
\label{future-work}

\subsubsection{Measuring curation activities}
\label{measuring-curation-activities}
Our analysis in Section~\ref{jira-ticket-corpus} focuses exclusively on the work log portion of the Jira ticket. We plan to incorporate other parts of the ticket such as the comments, which describe curatorial decisions in greater depth and incorporate the voices of project supervisors and managers in addition to curation staff. Analysis of the amount of time logged in Jira tickets will also allow us to further understand the intensity of specific curatorial actions (e.g., examining the differences between time estimated and the actual time required for curating studies). We will also triangulate our findings from this study with analysis available from processing history files and through qualitative interviews with curators. Processing history files are internal documents that contain commented syntax with statistical commands used to transform the deposited data files. They document all work done and changes made to deposited data, supporting reproducibility for data curation as new waves of data are added to a study. These files are referenced in work logs and provide more granular information about specific actions including \textit{data transformations} and \textit{documentation} steps. This will provide richer detail about how work is coordinated within the curation unit and by staff outside of it.  

Adding a second level of granularity to our schema will allow us to detect and differentiate specific curation activities. This will involve a second round of annotation with a version of the more detailed schema in Section~\ref{annotation-schema}. We are also interested in understanding typical sequences of curation actions in workflows and which actions tend to co-occur. Our computational model in Section~\ref{classifier} incorporates both individual words and bigrams, preserving common sequences of terms in work logs rather than modeling work log text as an unstructured bag of words. To improve the classifier, we will account for the order in which curation activities tend to occur; including factors for order of operations could address some of the confusion exhibited by the classifier. For example, \textit{initial review and planning} tends to happen at the beginning of curation work while \textit{quality checks} occur prior to and immediately after release of a study.

We will also compare frequent curation actions by archives to characterize the impacts that standardization efforts have had on curation work at ICPSR over the past several years (e.g., how adopted or mandated curation practices have percolated through the organization). Identifying curation work is a first step towards analyzing the relationship between data curation and use. The larger goal of our research is to connect curation activities with measures of data use and impact, including trust in data and in the archive itself \cite{Plantin2019-ba, Plantin2021-hl}.

\subsubsection{Understanding the impact of organizational structure on curation work}
\label{understanding-impact-of-organizational-structure}
We believe our analysis has revealed the continued impact of legacy organizational structures on curatorial work. ICPSR moved curation work out of topical archives and into a central unit in 2017, which changed the relationships between curators, archives, and data; the organization then adopted Jira to track and manage curation requests and work. The work logs available for analysis did not extend far enough back to reveal traces of ICPSR’s prior organizational structure and heterogeneity between topical archives. However, we did see evidence of evolving work practices following the centralization of the Curation Unit, the adoption of Jira, and standardized levels of curation. During the initial months of Jira use, curators broke curation requests into multiple tickets, with one ticket for each phase of curation, while in later years, each curation request was a single ticket that covered all phases.  

Additional context about Jira implementation, including reporting artifacts, is needed to interpret our findings. \textit{Data transformations} likely include work that curators do to make data compatible with ICPSR’s tools and methods; in such cases, there is effort spent that is unique to the institution and indirectly about improving the data. Curators must also track work by projects, and so they use Jira tickets to indicate all hours worked, which include non-curation activities. While we removed non-curation actions from our analysis, they accounted for a sizable portion of logged actions. This high level of non-curation activity may not be reported in the same way in other systems where curators are not required to keep track of their work hours in this way. Clerical work and other miscellaneous tasks are a necessary part of daily work in a large organization, but capturing these as part of a curation workflow obscures some of the higher value efforts made to improve the quality of deposited data. We also recognize that the transcripts curators generate in Jira are public within the organization; that means that the comments of subordinates (e.g., curators) are visible to those with power (e.g., supervisors, directors). Prior work acknowledges that “what can be part of any public transcript is also a matter of struggle” \cite{Gal1995-si}; we must remember that some comments or work descriptions will be purposefully vague or missing in order to resist preservation, domination, or other forms of control. Plantin \cite{Plantin2021-hl} argues that data processors use micro-resistance such as socializing and communicating expertise to avoid the alienation that can accompany their work. Future analysis must take into account the power dynamics at play in creating this documentation. 

Prior studies of issue tracking systems like Jira have mitigated similar issues by narrowly constraining their analysis to issue resolution time or by extracting sentiments from developer discourse \cite{Murgia2014-jd, Ortu2015-da}. The detection of non-curation actions, however, gives a more complete picture of a curator participating in the wide range of activities that an organization values but that extend beyond traditional curation work (e.g, professional development). ICPSR’s practice of capturing these activities, and our model’s ability to detect them, raise larger questions about what it means to meaningfully engage human curators in an increasingly standardized curation pipeline. Activities like professional development that are important to archives and their employees are not readily captured by taxonomies of curatorial action, and it will continue to be important to account for these kinds of resources as we support the humans in the data curation loop. Any model that doesn't account for these activities is missing a critical part of what it means to be an employee in any organization.

\subsubsection{Characterizing data curation – it’s not always a pipeline}
\label{characterizing-data-curation}
Our itemized approach to detecting curation activities fits a narrative in which the data processing pipeline is theorized as a factory-like workflow \cite{Plantin2021-hl}. In interrogating the limits of our own approach, we ask what a system like Jira might afford data workers in resisting both invisibility and accounting. One possible example of passive resistance to such invisibility is the variety of ways the Jira system is used. Our analysis of ticket length and complexity of syntax showed differences in the ways that curators used the Jira system. The amount of detail that curators included in their entries, approximated by the length of work log entries, varied from the detailed to the minimal or vague (e.g., "worked on curating study"). This also suggests variability in the value individual curators place on work logs; some may view work logs as administrative busywork and therefore document the bare minimum while others may incorporate work logs into their personal task management practices, seeing a benefit to adding more detail. Worklogs make curators’ work more visible, make it possible to acknowledge contributions, and capture divisions of labor: these goals are in line with Plantin’s \cite{Plantin2021-hl} steps to the emancipation of data workers. Balancing the visibility and exposure that work logs create is a critical challenge for organizations that employ them. We look forward to further exploring the benefits and risks of rendering curatorial work visible.

\section{Conclusion}
Our annotation schema suggests that curation work is not limited to data transformation or adequately captured by pipeline analogies. Instead, curatorial actions include many types of work such as communication and documentation that have not been effectively captured in prior descriptions. Our computational model identifies myriad curation actions and provides a mechanism for measuring their frequency and duration. By automating the analysis of curation work logs, we enable research that studies evolving curation activities. We illustrate the kinds of insights our model can reveal about curation work. Many of the activities that are core to the curators’ work (e.g., communicating about a study, reviewing curation plans) do not fit neatly into pipeline metaphors or narrow depictions of curation work. Instead, we show that planning, communication, and quality review are central to the curation work of data archives.

\section{Acknowledgment}
\label{acknowledgement}
We thank ICPSR curation supervisors including Rujuta Umarji, Julie Eady, Sharvetta Sylvester, Sara Del Norte, Lindsay Blankenship, Katey Pillars, Meghan Jacobs, and Scott Liening who provided feedback on our schema of curatorial actions. We also thank Amy Pienta (ICPSR) and Jeremy York (UMSI) for their comments on earlier drafts. This material is based upon work supported by the National Science Foundation under grant 1930645. This project was made possible in part by the Institute of Museum and Library Services LG-37-19-0134-19.

\section{Author Contributions}
\label{author-contributions}
Conceptualization, L.H., A.T., and D.A.; Methodology, S.L., L.H., A.T., D.A., and D.B.; Resources, D.B.; Data Curation, D.B.; Writing - Original Draft, S.L., A.T., L.H., D.A., and D.B.; Supervision, A.T. and L.H.; Project Administration, L.H. and D.B.

% \bibliographystyle{IEEEtran}
% \bibliography{references}

\end{document}